\documentclass[letterpaper,10pt,twocolumn]{article}
\usepackage[T1]{fontenc}
\usepackage[utf8]{inputenc}
\usepackage[margin=0.75in,columnsep=0.25in]{geometry}
\usepackage{cite}
\usepackage{amsmath,amssymb,amsfonts}
\usepackage{algorithmic}
\usepackage{graphicx}
\usepackage{textcomp}
\usepackage{xcolor}
\usepackage{balance} 
\usepackage{lmodern}
\usepackage{booktabs}
\usepackage{tcolorbox}
\usepackage{tabularx}
\usepackage{subcaption}
\usepackage{eso-pic}
\usepackage{multirow}
\usepackage{caption}
\usepackage{array}
\usepackage{url}
\usepackage[hidelinks]{hyperref}
\hypersetup{
  pdftitle={Rhythmic-Semantic Co-Speech Gestures for Emotional Dialogue in Robots},
  pdfauthor={Christian Arzate Cruz, Edwin C. Montiel-Vázquez, Stefanos Gkikas, Miguel Mejia Tobar, Randy Gomez},
  pdfsubject={Extended preprint of the shorter paper accepted at ACM HAI 2026}
}

\def\BibTeX{{\rm B\kern-.05em{\sc i\kern-.025em b}\kern-.08em
    T\kern-.1667em\lower.7ex\hbox{E}\kern-.125emX}}
\begin{document}


\title{Rhythmic-Semantic Co-Speech Gestures for Emotional Dialogue in Robots\thanks{Extended version of the shorter paper accepted at ACM HAI 2026 (\href{https://doi.org/10.1145/3841580.3845664}{doi:10.1145/3841580.3845664}). HAI peer review and acceptance apply only to the shorter paper.}}

\author{\parbox{0.97\textwidth}{\centering
Christian Arzate Cruz\textsuperscript{1},
Edwin C. Montiel-V\'azquez\textsuperscript{2},
Stefanos Gkikas\textsuperscript{1},\\
Miguel Mejia Tobar\textsuperscript{1},
Randy Gomez\textsuperscript{1}\\[0.6em]
\small
\textsuperscript{1}Honda Research Institute Japan, Wako, Saitama, Japan\\
\textsuperscript{2}School of Engineering and Sciences, Tecnologico de Monterrey,\\
Monterrey, Mexico\\[0.4em]
\footnotesize
\nolinkurl{christian.arzate@jp.honda-ri.com},
\nolinkurl{edwincmv@exatec.tec.mx},\\
\nolinkurl{stefanos.gkikas@jp.honda-ri.com},
\nolinkurl{miguel.mejiatobar@jp.honda-ri.com},
\nolinkurl{r.gomez@jp.honda-ri.com}
}}
\date{\small Extended preprint}

\maketitle

\begin{abstract}
Co-speech gestures can increase engagement and improve speech understanding. Yet, most data-driven robot systems generate either rhythmic beat-like motion or semantically meaningful iconic gestures, but not both. We present an emotion-aware rhythmic--semantic co-speech gesture system that generates iconic and beat behaviors in real time from text and the robot's internal emotional state. We formalize the task of beat--iconic gesture generation using emotion and propose lightweight models that predict iconic timing and intensity, then fill the remaining spans with emotion-conditioned beats at an appropriate intensity level. We train on the BEAT2 dataset and deploy the system on the social robot Haru. In an online user study with $60$ participants, our method shows stronger speech--gesture alignment and is preferred over baseline methods.
\end{abstract}

\section{Introduction}
\begin{figure}
    \centering
    \includegraphics[width=1.0\linewidth]{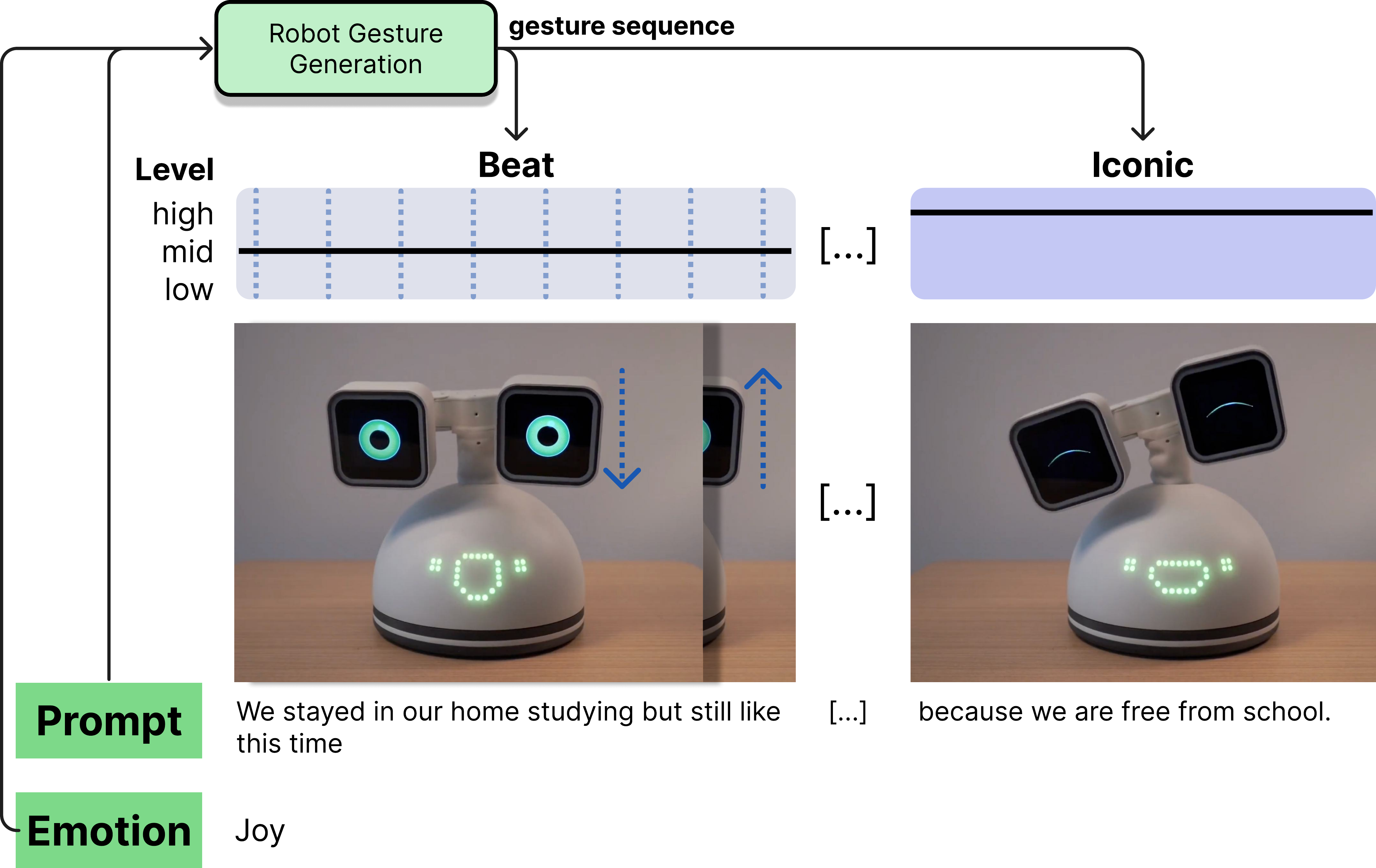}
    \caption{Pipeline overview. Given a text prompt and a target emotion (e.g., joy), the system generates a single co-speech gesture sequence that combines iconic and beat gestures with emotion-dependent intensity.}
    \label{fig:what}
\end{figure}

Humans convey meaning and internal states through body gestures and facial expressions. These cues range from large, deliberate movements that depict what people say to small, rhythmic motions that accompany speech. We call the former \emph{iconic} gestures and the latter \emph{beat} gestures~\cite{nyatsanga2023comprehensive}. Prior work on robot co-speech gestures rarely models both types in one system, unlike recent state-of-the-art (SOTA) co-speech generation systems for artificial agents~\cite{liu2024emage, zhang2025semtalk}. Therefore, we aim to generate a single co-speech gesture system that includes both iconic and beat gestures, while remaining coherent with the utterance and the robot's emotional state (Figure~\ref{fig:what}).

Robots must also adapt motion to emotion. Many co-speech approaches do not explicitly model how emotion changes movement~\cite{neff2008gesture, kucherenko2020gesticulator, wu2021probabilistic}. We focus on four emotions from Plutchik's wheel~\cite{Plutchik2001nature}: joy, anger, sadness, and fear. Given an utterance and a target emotion, our system generates co-speech gestures that communicate both \emph{what} the robot says and \emph{how} it feels. In this paper, we use the term \emph{iconic} to denote content-aligned co-speech animations that convey the intended \emph{emotional meaning} of what is being said (i.e., emotion-specific variants), while \emph{beat} gestures refer to non-semantic rhythmic motions that accompany speech. 

Another key requirement in robotics is responsiveness. Many data-driven methods assume speech audio is available at inference time for prosody or synchronization~\cite{kucherenko2020gesticulator,yoon2019robots,li2024learning,fernandez2025evaluating,wu2021probabilistic, neff2008gesture}. This assumption can add latency for interactive robots that use expressive text-to-speech (TTS) (e.g., about 2\,s with GPT-SoVITS~\cite{gpt_sovits_github}). We therefore design a text-only system that takes only the utterance and the intended emotion.

Based on these constraints, we formalize a new robotics task: efficient, text-only generation of joint beat--iconic co-speech gestures that vary with emotion. We train lightweight models on the \textbf{b}ody-\textbf{e}xpression-\textbf{a}udio-\textbf{t}ext (BEAT2) dataset~\cite{liu2024emage} and generate iconic and beat motion in one coherent sequence. We deploy the system on the social robot Haru~\cite{Gomez2018haru,Arzate2025_hri,Arzate2025_roman,Arzate2024} and evaluate it in an online user study ($N=60$). Participants rate our method higher for speech--gesture correlation across emotions and prefer it in most conversations compared to baselines.

Our contributions are as follows:

\textbf{1)} We define a new task for social robots: generating a single co-speech gesture stream that fuses beat and iconic gestures while accounting for emotion. This allows the use of BEAT2 for non-humanoid robots.

\textbf{2)} We propose lightweight models that predict which words trigger iconic gestures and estimate their intensity only from text and emotion. Likewise, we train the same model architectures to predict beat gesture intensity.

\textbf{3)} We integrate all our models into a gesture generation system that runs in real time without audio at inference and can map to different non-humanoid robot embodiments.

\section{Related Work} \label{sec:related_work}
In this section, we review prior work on co-speech gesture generation for robots and artificial agents.

\subsection{Robots}
In robotics, few approaches use unique strategies for generating co-speech motion, such as retargeting motion from recorded human videos to a robot~\cite{go2018andorid} or rule-based policies for beat or hand-gestures~\cite{bremner2009beat}. In contrast, most recent research focuses on generating co-speech gestures for the upper body of humanoid robots using data-driven learning~\cite{li2024learning}. These works typically collect paired speech and motion from people delivering monologues or speeches, use audio and text as input, and optimize losses that encourage temporally coherent head and arm motion that matches speech rhythm. Methods differ mainly on how linguistic and prosodic information is represented and the learning architecture used.

With respect to linguistic features, some approaches incorporate sentence-level context using embeddings~\cite{kucherenko2020gesticulator} from the Bidirectional Encoder Representations from Transformers (BERT) language model~\cite{devlin2019bert}, while others use word-level embeddings to provide finer-grained lexical conditioning~\cite{yoon2019robots}. With respect to model architectures, many works formulate co-speech gesture generation as a regression problem over joint poses, commonly using autoregressive models that condition on previously generated poses to maintain motion continuity~\cite{kucherenko2020gesticulator,yoon2019robots,li2024learning,fernandez2025evaluating}. Other approaches use generative adversarial networks (GANs) to better match the distribution of human motion and reduce over-smoothing~\cite{liu2023speech, yu2020srg}.

A key limitation of current robotics approaches is that they rarely combine iconic and beat gestures in one system. Most learned generators focus on beat-like co-speech motion~\cite{kucherenko2020gesticulator, yoon2019robots, li2024learning, liu2023speech,yu2020srg}, while the few that target iconic gestures often select or trigger pre-defined animations without modeling their timing~\cite{fernandez2025evaluating}. In our work, we generate both in a single pipeline. First, an \textit{Iconic Module} predicts when to trigger content-aligned emotional animations and how strongly to perform them, conditioned on linguistic context and the target emotion. Second, a \textit{Beat Module} fills the remaining spans with rhythmic gestures by predicting emotion-conditioned beat activation intensities. To the best of our knowledge, this is the first robot co-speech architecture that explicitly models both iconic and beat gestures.

\subsection{Artificial Agents}
Co-speech gesture generation for artificial agents has received more attention than robotics, largely due to the availability of large datasets. A widely used resource is BEAT and its extension BEAT2~\cite{liu2024emage}, which pair motion capture with audio and text, and include word-level iconic intensity annotations. We use BEAT2 to train both our iconic and beat components because it provides strong supervision for joint rhythmic and semantic co-speech modeling.

Recent agent systems move beyond upper-body gestures~\cite{wu2021probabilistic, neff2008gesture} toward full-body motion~\cite{gao2025sarges}, improved speech synchronization~\cite{pang2025llm}, and more diverse motion~\cite{liu2024towards}. These methods typically target realistic humanoid characters and often condition on speech audio at inference. In contrast, we formalize a robotics task with stricter constraints.

Most agent approaches use deep sequence models (e.g., LSTMs~\cite{Hochreiter1997long}, transformers~\cite{vaswani2017attention}) and sometimes diffusion or adversarial objectives to improve realism and diversity~\cite{liu2024emage, yang2023diffusestylegesture}. Some recent work also uses LLMs to guide motion or label generation~\cite{chen2025motion, gao2025sarges}. While we also follow a data-driven approach, our focus is not photorealistic motion quality but deployable robot behavior under latency constraints. We therefore avoid audio at inference and use lightweight predictors for iconic gestures and beat intensity, conditioned on emotion and text.


Finally, prior work explores stylized character animation, such as exaggerated facial expressions paired with beat gestures for non-photorealistic agents~\cite{omine2025co}. This direction relates to our setting because non-humanoid robots often benefit from hand-designed iconic content that fits their embodiment. Our approach differs in focus: rather than only synthesizing motion, we predict iconic placement and intensity within the utterance and generate the surrounding beat behavior in a unified system.

\section{Task Overview} \label{sec:system_overview}
\begin{figure}[tb]
    \centering
    \includegraphics[width=0.95\linewidth]{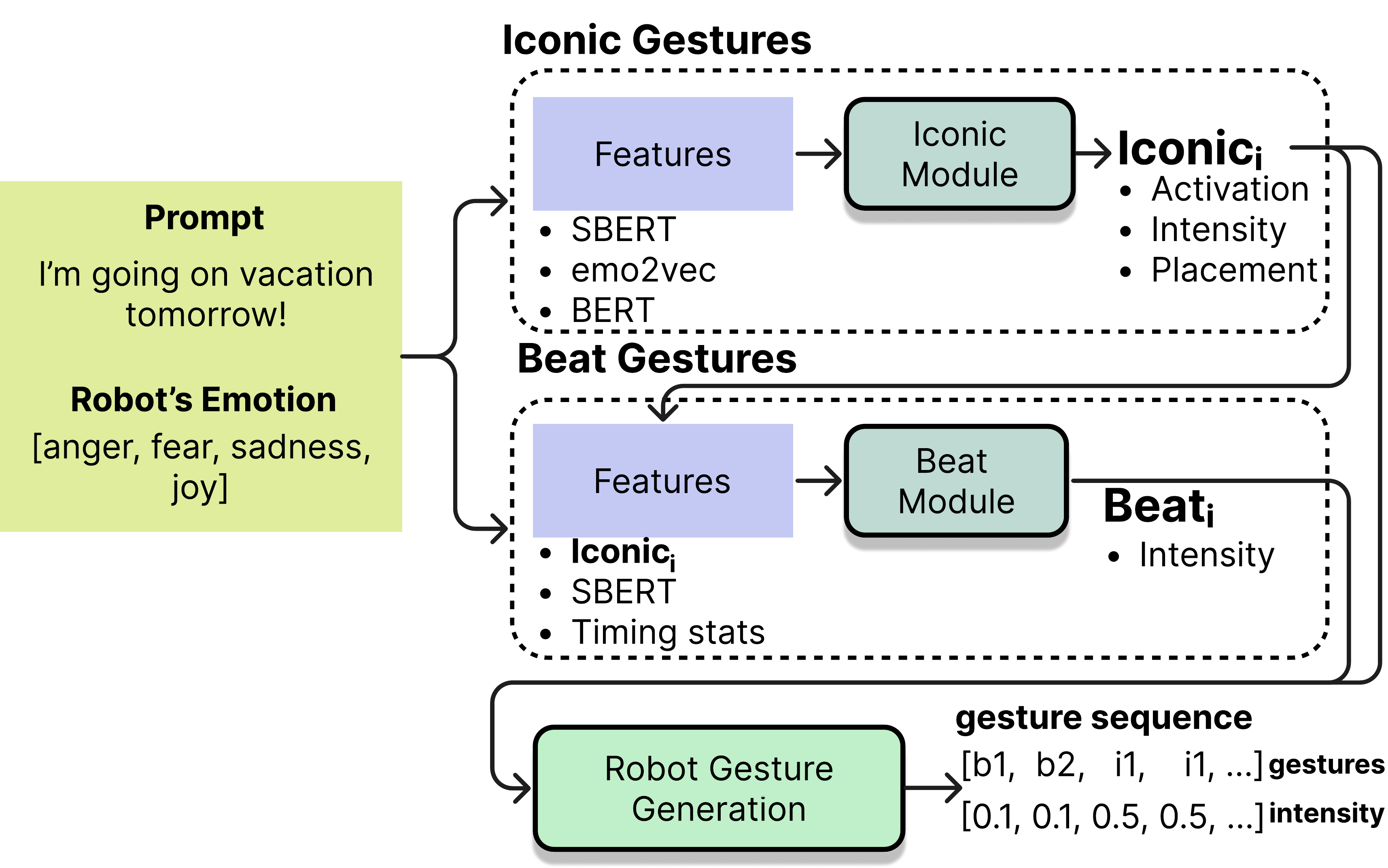}
    \caption{Rhythmic--semantic co-speech gesture system overview.}
    \label{fig:overview}
\end{figure}

Figure~\ref{fig:overview} summarizes our rhythmic--semantic co-speech pipeline. At inference, the system takes a text prompt and a target emotion and outputs a time-aligned gesture sequence that combines iconic (emotion-meaning) and rhythmic (beat) behaviors.

For the iconic sub-task, we encode the utterance with BERT~\cite{devlin2019bert} or SBERT~\cite{reimers2019sentence} and emo2vec~\cite{xu2018emo2vec}, depending on the model. These embeddings are passed to the \textit{Iconic Module}, which predicts (i) a binary sentence-level activation flag, (ii) an intensity value in $[0,1]$, and (iii) a placement as a word index. We represent these outputs as $Iconic_i$, obtained using a classifier (activation) and a regressor (intensity and placement).

A key element of our approach is to explicitly condition our models on the target emotion. BEAT2 provides emotion labels, which we use to define the four basic emotion classes in our study: joy, sadness, anger, and fear.

For beat gestures, BEAT2 provides no intensity annotations, so we derive proxy beat intensity from motion derivatives in spans without iconic gestures and aggregate it over fixed windows. Since robots support only a few perceptually distinct beat amplitudes, we discretize this proxy into levels and train the \textit{Beat Module} as a classifier. The \textit{Beat Module} uses SBERT~\cite{reimers2019sentence} and sentence-level timing features (e.g., number of words, mean word duration, and utterance duration) to predict beat intensity for all windows $t$ while avoiding overlap with iconic spans ($Beat_t$). Finally, we merge $Iconic_i$ and $Beat_t$ and execute the resulting sequence through the robot controller. The following sections detail training and evaluation using BEAT2~\cite{liu2024emage}.

\section{Iconic Gestures} \label{sec:iconics}
In this section, we describe the iconic gesture sub-task, which our \textit{Iconic Module} handles (see Figure~\ref{fig:overview}). The module includes two models: a sentence-level \textit{iconic classifier} and a word-level \textit{intensity regressor}.

\subsection{Inputs and Targets}
The two main inputs to our system, for all models, are the utterance the robot will say and the robot's internal emotional state. For training, we obtain both from BEAT2~\cite{liu2024emage}. BEAT2 provides word-level iconic annotations as continuous intensity values. We use these labels in two ways: (1) as the regression target for iconic intensity, where the peak also indicates placement, and (2) to derive a binary activation label that indicates whether the utterance contains an iconic.

We set the binary label to $1$ if any word in the utterance has iconic intensity $>0$. We use this threshold because iconic activations are sparse in BEAT2. The dataset contains an average of $0.87$ iconic gestures per utterance (counting contiguous nonzero segments as one gesture). Therefore, we model iconic activation as a sentence-level binary classification task.

\subsection{Models} \label{subsec::iconic_models}
We explore two model families: an LLM-driven approach and data-driven models. Specifically, we evaluate GPT-4o~\cite{openai_gpt-4_2024}, BERT~\cite{devlin2019bert}, and a lightweight Transformer (TRS)~\cite{vaswani2017attention}. We use these models for the different sub-tasks in our gesture generation pipeline.

TRS is an efficient transformer variant that uses cross- and self-attention with a compact latent space. Instead of applying attention directly over all input embeddings, it introduces a compact latent space as an intermediate representation and attends through these latent tokens. This design reduces computation while preserving global context. We select the TRS implementation in \cite{gkikas2025efficient} for its efficiency and low resource requirements and use it for both classification and regression.

For the \textit{iconic classifier} and \textit{intensity regressor}, we use prompting for the LLM-driven baseline and train the data-driven models on BEAT2. This setup lets us compare lightweight models against an SOTA LLM that can capture richer semantic cues~\cite{havlik2024meaning}.

\subsection{Training}
We segment the original dataset transcripts into sentences of at most $40$ words using an LLM. We prompt the model to recover complete sentence boundaries from the BEAT2 monologues, which contain no punctuation. The mean length throughout the dataset was $29$ words, with a median of $18$. Filtering led to a mean closer to the median ($20$), while affecting less than $1\%$ of the dataset. As such, the resulting dataset was around $17000$ samples. We obtain training and testing sets using an $80/20$ random split. However, we split the data at the full utterance level, rather than the sentence-level, to ensure that the dataset remained balanced. This data is used to train both data-driven models until no improvement was observed. 

For the BERT classifier, we separate the text and the emotion using a special token, then apply the default embeddings~\cite{devlin2019bert}. The expression for this is: $i_n = BERT_{embed}(s_n +   [SEP] + e_n)$, where $s_n= (w_1, w_2, \ldots, w_{40})$ represents the sentence, and $e_n$ the emotion. Meanwhile, the BERT intensity regressor uses a default per-token regression configuration.

TRS uses the smaller sentence-level SBERT embedding layer~\cite{reimers2019sentence} for the classification task. Meanwhile, it performs regression using individual sentence and word-of-interest pairs $p_{n,k} = (s_n,w_k)$, $k \in [1,40]$, where $s_n$ represents the sentence and $w_k$ represents an individual word, effectively highlighting the given word in $s_n$. The sentence is encoded using SBERT~\cite{reimers2019sentence}, while $w_k$ is embedded using emo2vec~\cite{xu2018emo2vec} averaged with the representation of the emotion label to include affective information~\cite{zhang2025semtalk}. 

The LLM-based approach uses two prompts to (i) predict whether an utterance contains an iconic gesture (binary label) and (ii) estimate a per-word intensity sequence. Both prompts include short definitions of iconic and beat gestures to guide the model.
 
\subsection{Evaluation}
We evaluate the iconic classifier and regressor using automatic metrics. For classification, we report macro-averaged accuracy (acc), precision (p), recall (r), and F1 score (F1). For regression, we report mean absolute error (MAE), root mean squared error (RMSE), and Spearman correlation. We also report the average inference time, including embedding, in milliseconds (t). These metrics evaluate the \emph{iconic activation} and \emph{iconic intensity}.

We also evaluate \emph{iconic placement} as a span-matching task. Since an iconic spans multiple words, we extract the peak word (highest intensity) from the prediction. Then, we count a placement as correct if the predicted peak falls within a 4.5\,s window of a ground truth activation. We choose 4.5\,s because it matches the mean duration of Haru's iconic animations, which expert animators designed to express emotions. We also report iconic placement accuracy using the average activation time in the dataset: 1.16s.

Table~\ref{tab:iconic_all_results} summarizes results for \textit{Iconic Activation}, \textit{Iconic Placement}, and \textit{Iconic Intensity} across architectures. For placement, we compute accuracy only on test samples that contain an iconic gesture in the ground-truth. Overall, BERT achieves the strongest performance across sub-tasks.

\begin{table*}[ht]
  \centering
  \caption{Results for iconic sub-tasks. Underlined values are best per column.}
  \label{tab:iconic_all_results}
  \footnotesize
  \setlength{\tabcolsep}{3pt}
  \renewcommand{\arraystretch}{1.2}

  \begin{tabular}{lccccc|cccc|cc}
    \toprule
    & \multicolumn{5}{c|}{\textbf{Iconic Activation}} &
      \multicolumn{4}{c|}{\textbf{Iconic Intensity}} &
      \multicolumn{2}{c}{\textbf{Iconic Placement}} \\
    \cmidrule(lr){2-6}\cmidrule(lr){7-10}\cmidrule(lr){11-12}
    \textbf{Model} &
    \textbf{Acc} $\Uparrow$ & \textbf{P} $\Uparrow$ & \textbf{R} $\Uparrow$ & \textbf{F1} $\Uparrow$ & \textbf{t} $\Downarrow$ &
    \textbf{MAE} $\Downarrow$ & \textbf{RMSE} $\Downarrow$ & \textbf{Spearman} $\Uparrow$ & \textbf{t} $\Downarrow$ &
    \textbf{Acc(4.5s)} $\Uparrow$ &
    \textbf{Acc(1.16s)} $\Uparrow$\\
    \midrule
    \textbf{LLM}  & 0.5078 & 0.5976 & 0.5078 & 0.4013 & 596.369 & 0.0824 & 0.2206 & 0.0633 & 596.369 & 0.6582 & 0.3772 \\
    \textbf{TRS}  & 0.5910 & 0.5878 & 0.5910 & 0.5865 & 4.243   & 0.0790 & 0.1525 & 0.1572 & 56.298  & 0.6228 & 0.4371\\
    \textbf{BERT} & \underline{0.6095} & \underline{0.6134} & \underline{0.6095} & \underline{0.6106} & \underline{3.733} & 
    \underline{0.0595} & \underline{0.1436} & \underline{0.1946} & \underline{4.209} &
    \underline{0.6639} & \underline{0.4971} \\
    \bottomrule
  \end{tabular}
\end{table*}

\section{Beat Gestures Generation} \label{sec:method_beats}
This section describes the rhythmic (beat-like) gesture generation sub-task for the \textit{Beat Module} (see Figure~\ref{fig:overview}), which predicts the sentence-level intensity of beat gestures conditioned on emotion and the predicted placement of iconic gestures. Prior analyses of BEAT2 suggest that emotion primarily modulates the \emph{amount} of motion (activation and intensity) rather than distinct beat \emph{types}~\cite{liu2024emage}. We therefore focus on predicting beat intensity while avoiding interference with iconic gestures.

We discretize each utterance into fixed windows of length $\Delta$ and treat each window as a unit opportunity (time to perform a beat gesture). We set $\Delta=0.8$\,s based on empirical robot tests: shorter windows yielded jittery motion, while longer windows reduced responsiveness. 

For each sentence $s$, we predict a beat intensity label $\{\texttt{low},\texttt{mid},\texttt{high}\}$. Given the predicted iconic spans, we mark any window that overlaps an iconic as \texttt{overlap} and exclude it from beat generation. At runtime, we apply $\ell(s)$ to all non-\texttt{overlap} windows within the sentence to obtain the window sequence $Beat_t$.

\subsection{Sentence-level Beat Intensity Labels} \label{sec:sentence_intensity_labels}
Since BEAT2 does not provide beat-gesture intensity annotations, we derive sentence-level beat intensity labels from motion in fixed windows. We mask all frames that overlap any word with iconic intensity $>0$; windows with no remaining frames are marked as \texttt{overlap} and excluded.

For each remaining window $t$, we compute a motion-energy score $E_t$ from first-order pose changes of the head and both arms, which show the clearest emotion-dependent differences. We then normalize this score into a window intensity:
\begin{equation*}
I_t \;=\; \min\!\left(1,\; \frac{\max(0,\,E_t-\tau)}{\kappa}\right),
\label{eq:window_intensity}
\end{equation*}
where $\tau$ is a training-set baseline and $\kappa$ is a training-set scale factor. In our implementation, we estimate a separate baseline per body part using a quantile rule (with target activity $\alpha=0.30$), and we set $\kappa$ to the 99th percentile of positive energies computed on the training split. We reuse these values for validation and test.

We aggregate window intensities to a sentence-level score by averaging over the set of non-\texttt{overlap} windows $\mathcal{T}(s)$ within sentence $s$:

\begin{equation*}
S(s) \;=\; \frac{1}{|\mathcal{T}(s)|}\sum_{t\in\mathcal{T}(s)} I_t,
\label{eq:sentence_score}
\end{equation*}

and discretize $S(s)$ into three ordinal classes $\ell(s)\in\{\texttt{low},\texttt{mid},\texttt{high}\}$ using training-set quantiles:
\begin{equation*}
\ell(s)=
\begin{cases}
\texttt{low}, & S(s)\le q_{0.33}\\
\texttt{mid}, & q_{0.33} < S(s)\le q_{0.66}\\
\texttt{high}, & S(s) > q_{0.66},
\end{cases}
\label{eq:sentence_label}
\end{equation*}
where $q_{0.33}$ and $q_{0.66}$ are computed on the training split and reused for validation/test. This quantile-based discretization avoids manual threshold tuning and yields balanced sentence-level supervision.

\subsubsection{Label agreement with speech rhythm cues}
We validate our motion-energy beat-intensity targets by correlating them with audio rhythm proxies (log-RMS energy and onset strength). We compute Spearman $\rho(\delta)=\rho(y_t,a_{t+\delta})$, where $\delta$ is a time shift applied to the audio window $[t_{\text{start}}+\delta,\,t_{\text{end}}+\delta]$. At $\delta=0$, correlations are positive (log-RMS: $\rho=0.203$; onset: $\rho=0.241$; $n=119{,}515$). Searching $\delta\in[-0.4,0.4]$\,s (0.02\,s step) yields a best alignment at $\delta=+0.4$\,s (log-RMS: $\rho=0.226$; onset: $\rho=0.264$; $n=119{,}461$), supporting that the proxy labels capture speech-rhythm structure with a systematic timing offset.

\subsection{Models}
We evaluate the same model families used for the iconic sub-task (Section~\ref{subsec::iconic_models}): an LLM, a lightweight transformer, and BERT-style encoders.

For the LLM baseline, we prompt the model to classify sentence-level beat intensity given the utterance and the target emotion. The prompt also includes brief definitions of iconic and beat gestures to guide the model. For encoder-based classifiers, we test BERT, as in Section~\ref{subsec::iconic_models} and also RoBERTa~\cite{liu2019roberta}. We include RoBERTa as beat intensity prediction depends on subtle linguistic and rhythm cues. RoBERTa uses the same encoder architecture as BERT with a stronger pretraining setup, and it often improves sentence classification performance as a drop-in replacement~\cite{liu2019roberta}.

For the BERT-based variants, the main input is the utterance transcript. We prepend a small set of control tokens that encode gesture-relevant context: the target emotion, iconic cues, and discretized timing descriptors. The iconic cues come from the Iconic Module output ($Iconic_i$), which indicates the predicted presence and intensity of gestures in the utterance. The timing descriptors include the number of words, utterance duration, and average word duration (ms). We do not use the ground-truth word timings in BEAT2, since they come from human recordings and are not available at inference on a robot. Instead, we estimate timing from text during both training and inference using a lightweight heuristic (no TTS synthesis): we assign word durations from token length, add fixed pauses for punctuation, rescale durations to match an emotion-dependent target speech rate, and accumulate them to obtain per-word start/end times and the total utterance duration.

For the transformer baseline, we use the same timing descriptors, a one-hot vector for the target emotion, and SBERT embeddings.

We summarize the results of all beat gesture generation models in Table~\ref{tab:beat_models_results}. RoBERTa achieves the best performance across all metrics, except inference time, where BERT is about $20\%$ faster.

\begin{table}[tbh]
  \centering
  \caption{Macro-averaged results for beat gesture intensity per sentence. Underlined values are best per column.}
  \label{tab:beat_models_results}
  \footnotesize
  \setlength{\tabcolsep}{5pt}
  \renewcommand{\arraystretch}{1.15}
  \begin{tabular}{lccccc}
    \toprule
    \multicolumn{6}{c}{\textbf{Beat Intensity}} \\
    \cmidrule(lr){1-6}
    \textbf{Model} & \textbf{Acc} $\Uparrow$ & \textbf{P} $\Uparrow$ & \textbf{R} $\Uparrow$ & \textbf{F1} $\Uparrow$ & \textbf{t} $\Downarrow$ \\
    \midrule
    \textbf{LLM}     & 0.3253 & 0.3263 & 0.3253 & 0.3253 & 961.307 \\
    \textbf{TRS}     & 0.3454 & 0.3791 & 0.3454 & 0.2997 & 4.259 \\
    \textbf{BERT}    & 0.3712 & 0.3506 & 0.3712 & 0.2874 & \underline{3.880} \\
    \textbf{RoBERTa} & \underline{0.4017} & \underline{0.4054} & \underline{0.4054} & \underline{0.4015} & 4.658 \\
    \bottomrule
  \end{tabular}
\end{table}

\section{Robot Gesture Generation} \label{sec:robot_gesture_generation}
We combine the best-performing models from each stage into a complete robot gesture generation system. It consists of four stages that run sequentially.

We start with (i) \textit{iconic classification}, which predicts whether a sentence should include an iconic gesture. For this stage, we use the BERT classifier. Next, we perform (ii) \textit{iconic placement} using the regression models to predict where in the sentence an iconic should start (the iconic peak). We select BERT for this stage as well, since it provides the strongest overall performance among the practical baselines. Using the same regressor output, we then assign (iii) \textit{iconic intensity} to each predicted iconic. These steps generate the vector $Iconic_i$ that marks when to trigger an iconic gesture, its intensity, and placement.

Finally, we estimate (iv) \textit{beat intensity} at the sentence level, using $Iconic_i$ as input. RoBERTa achieves the best performance in this sub-task, so we use it despite being slightly slower than BERT. This model outputs a window-level sequence $Beat_t$. Since our pipeline relies mainly on BERT-based components, we refer to it as \textbf{G-BERT} (\textbf{G}esture-conditioned \textbf{B}ERT). We refer readers to the supplementary video\footnote{\url{https://youtu.be/csBZXYNOr5w}}, which includes multiple examples of our implementation on Haru.

Combining $Iconic_i$ and $Beat_t$ yields a gesture sequence for a given utterance and target emotion. We map these predictions to executable robot motions by selecting and scheduling concrete animations based on the robot's embodiment and available expressive channels. In this section, we describe how we implement this mapping on the social robot Haru.

\subsection{Mapping $Iconic_i$ to Haru iconic gestures}
Haru executes iconic gestures from a library of hand-crafted animations designed to convey personality and emotion. We discretize the predicted iconic intensity into three levels (\textit{low}, \textit{medium}, \textit{high}), and for each emotion and intensity level we select a corresponding iconic animation aligned with Plutchik's wheel of emotions. This mapping was created together with a behavioral expert, who iteratively refined the gestures by reviewing videos of Haru across emotional contexts and interacting with the robot in person for approximately three hours.

For each of the four target emotions, we use three intensity levels that follow Plutchik's intensity scale: (\textit{serenity}, \textbf{joy}, \textit{ecstasy}), (\textit{annoyance}, \textbf{anger}, \textit{rage}), (\textit{pensiveness}, \textbf{sadness}, \textit{grief}), and (\textit{apprehension}, \textbf{fear}, and \textit{terror}). We show the iconic gestures for joy in Table~\ref{tab:joy_example} and all are present in the supplementary video.

\begin{table}[htb]
  \centering
  \small
  \caption{Example mapping for the \textbf{joy} iconic gestures.}
  \label{tab:joy_example}
  \setlength{\tabcolsep}{4pt}
  \renewcommand{\arraystretch}{1.10}
  \begin{tabular}{ccc}
    \toprule
    \textbf{Low} & \textbf{Medium} & \textbf{High} \\
    \midrule
    \textit{Serenity} & \textit{Joy} & \textit{Ecstasy} \\
    \midrule
    \includegraphics[width=0.29\columnwidth]{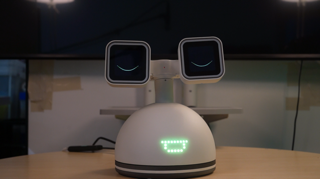} &
    \includegraphics[width=0.29\columnwidth]{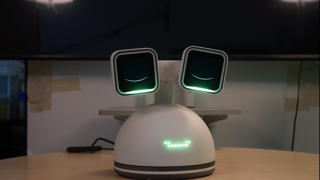} &
    \includegraphics[width=0.29\columnwidth]{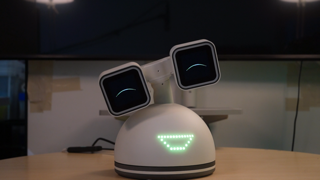} \\
    \bottomrule
  \end{tabular}
\end{table}

\subsection{Mapping $Beat_t$ to Haru beats}
Given the selected iconic and its timing (placement and duration), we pass this information, together with the remaining input features for RoBERTa, to the \textit{Beat Module}. The model predicts the beat intensity sequence $Beat_t$ while avoiding overlap with iconic spans. Since beat gestures do not need to convey specific semantics~\cite{liu2024emage}, we generate beats on the fly using Haru's available degrees of freedom.

We define ten beat templates that cover Haru's head and eye motion primitives (tilt, lean, eye rotation, eye glance, and base rotation). Each template has two directional variants (e.g., left/right, up/down, or clockwise/counter-clockwise), in Table~\ref{tab:beats_examples} we illustrate only one version. The ten templates are shown in the supplementary video. We map the normalized intensity in $Beat_t$ to motion amplitude by scaling between Haru's comfortable minimum and maximum ranges and speeds for each primitive. This enables smooth interpolation, where higher predicted intensity yields larger and faster beat motion. At each beat onset, we randomly select a template (and its direction) while preventing the same template from being selected twice in a row.

\begin{table}[htb]
  \centering
  \small
  \caption{Example of Haru's motion primitives for beat gesture.}
  \label{tab:beats_examples}
  \setlength{\tabcolsep}{4pt}
  \renewcommand{\arraystretch}{1.10}
  \begin{tabular}{ccccc}
    \toprule
    \textbf{Tilt} & \textbf{Glance} & \textbf{Lean} & \textbf{Eyes} & \textbf{Base} \\
    \midrule
    \includegraphics[width=0.15\columnwidth]{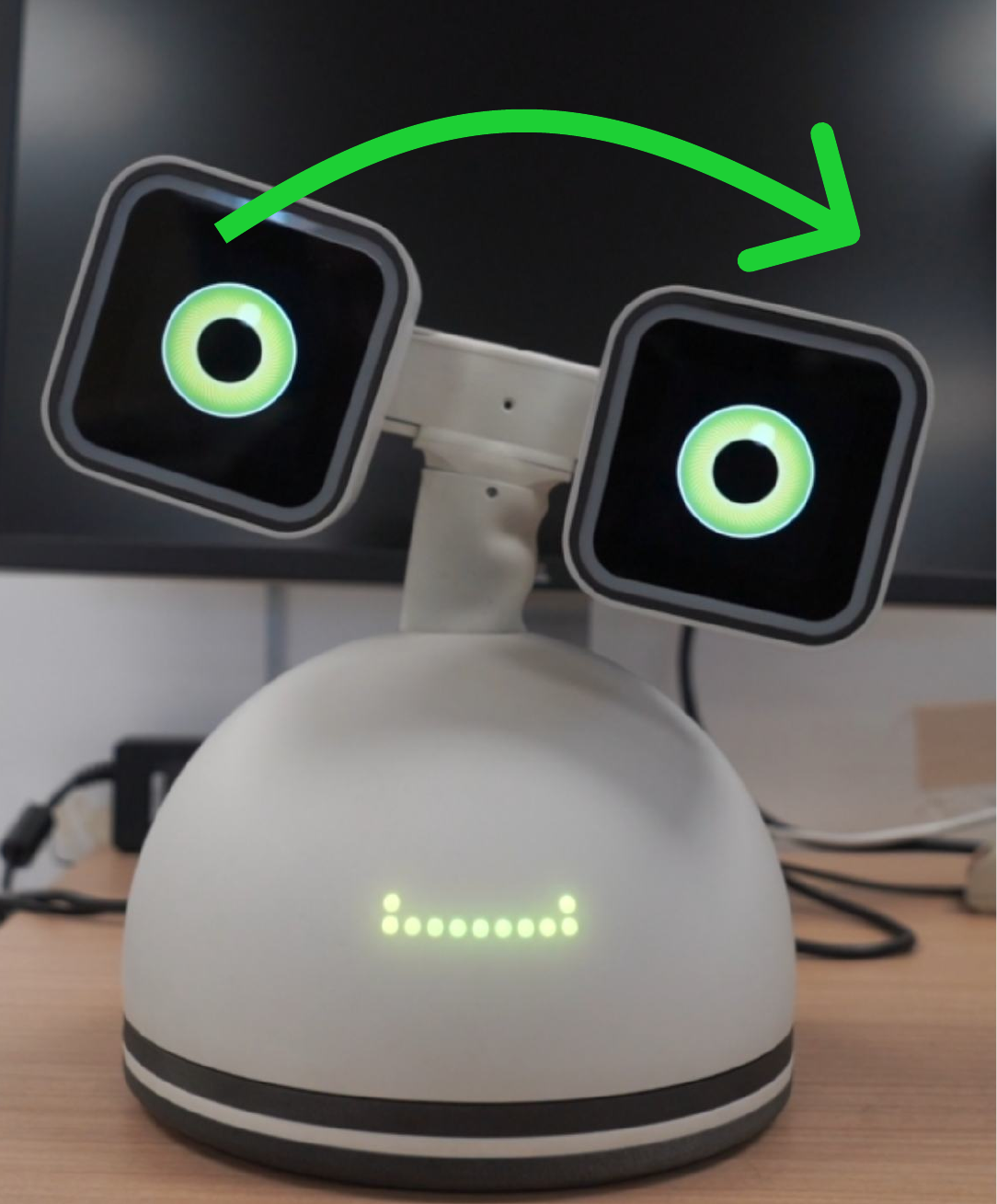} &
    \includegraphics[width=0.15\columnwidth]{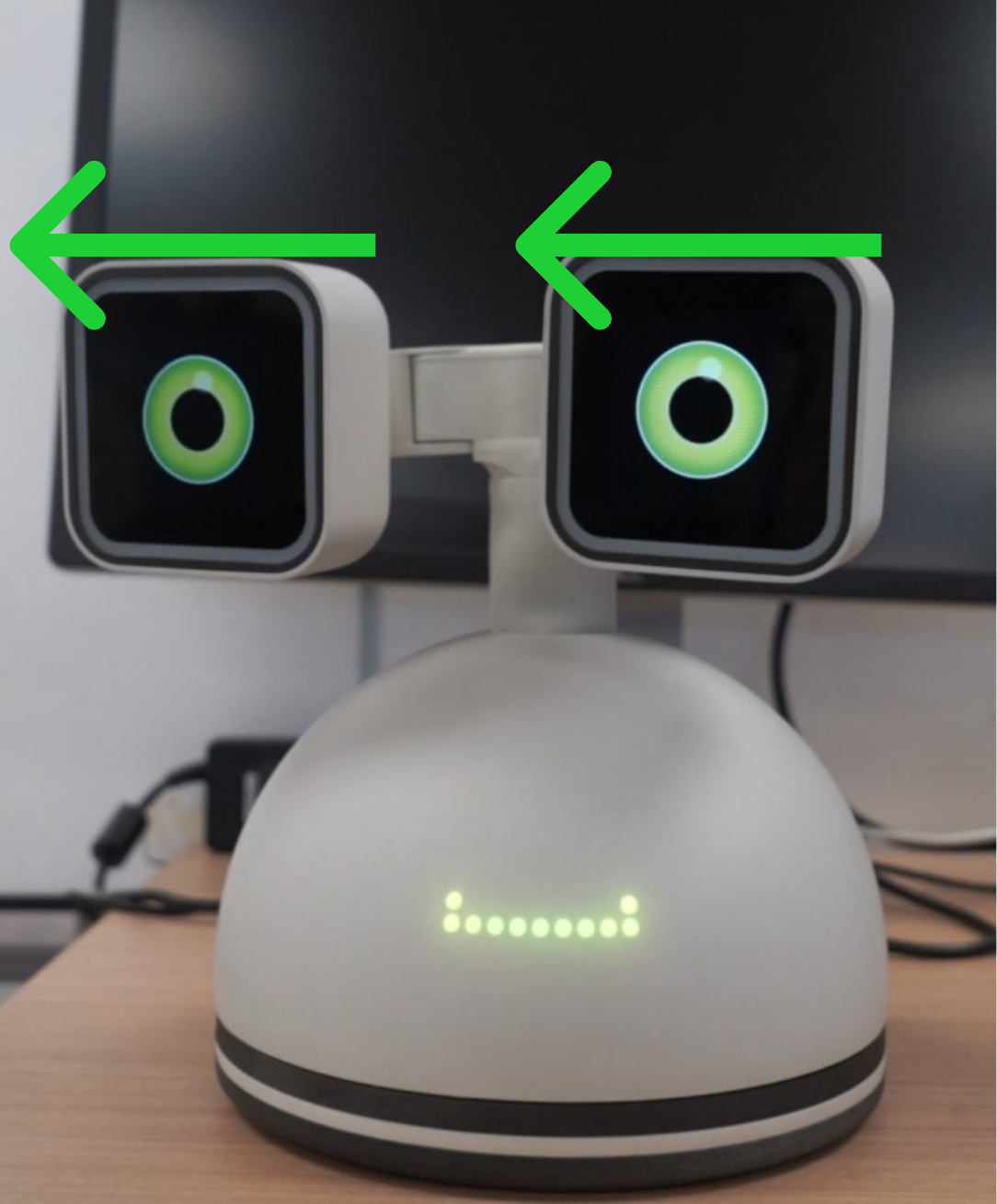} &
    \includegraphics[width=0.15\columnwidth]{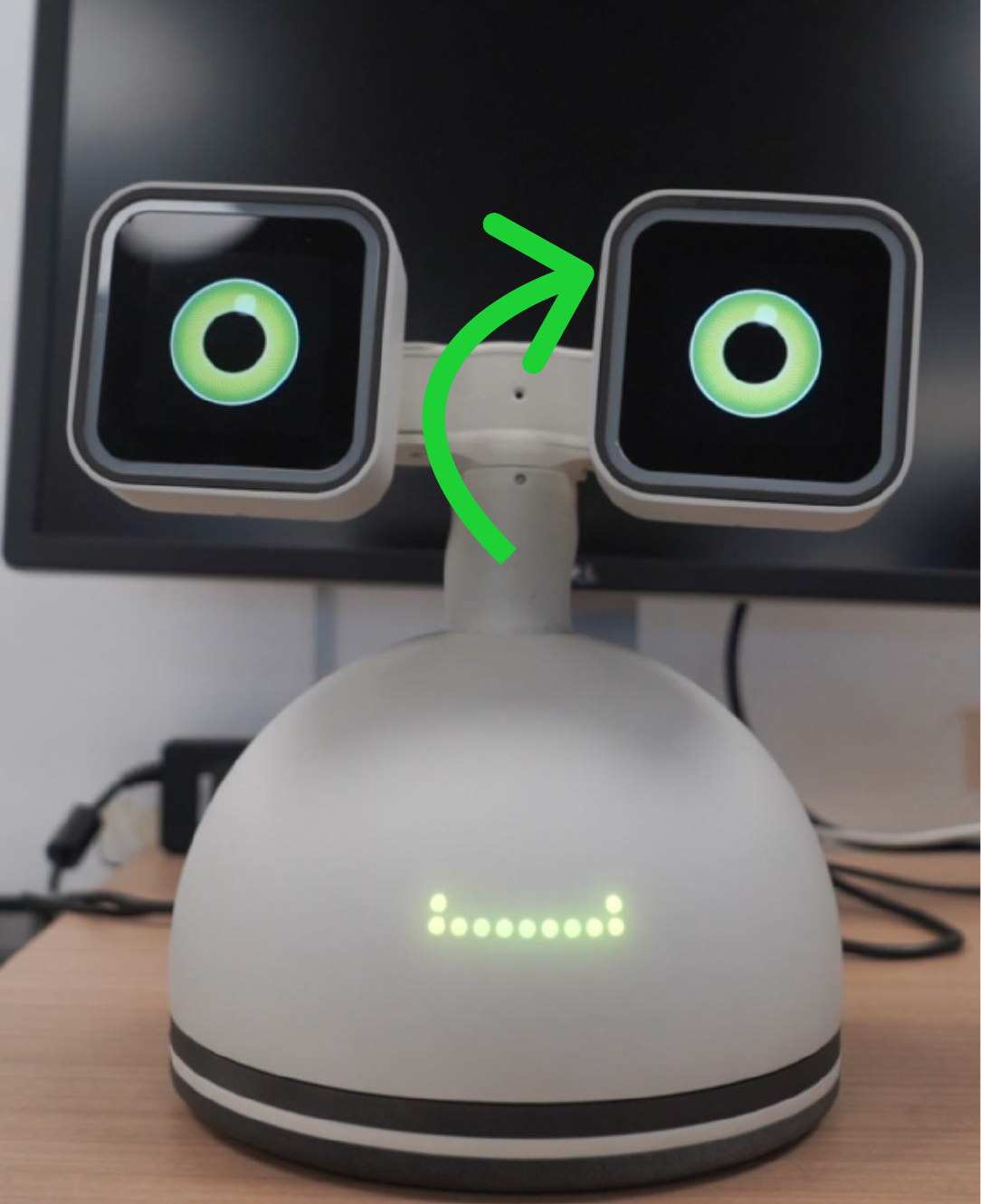} &
    \includegraphics[width=0.15\columnwidth]{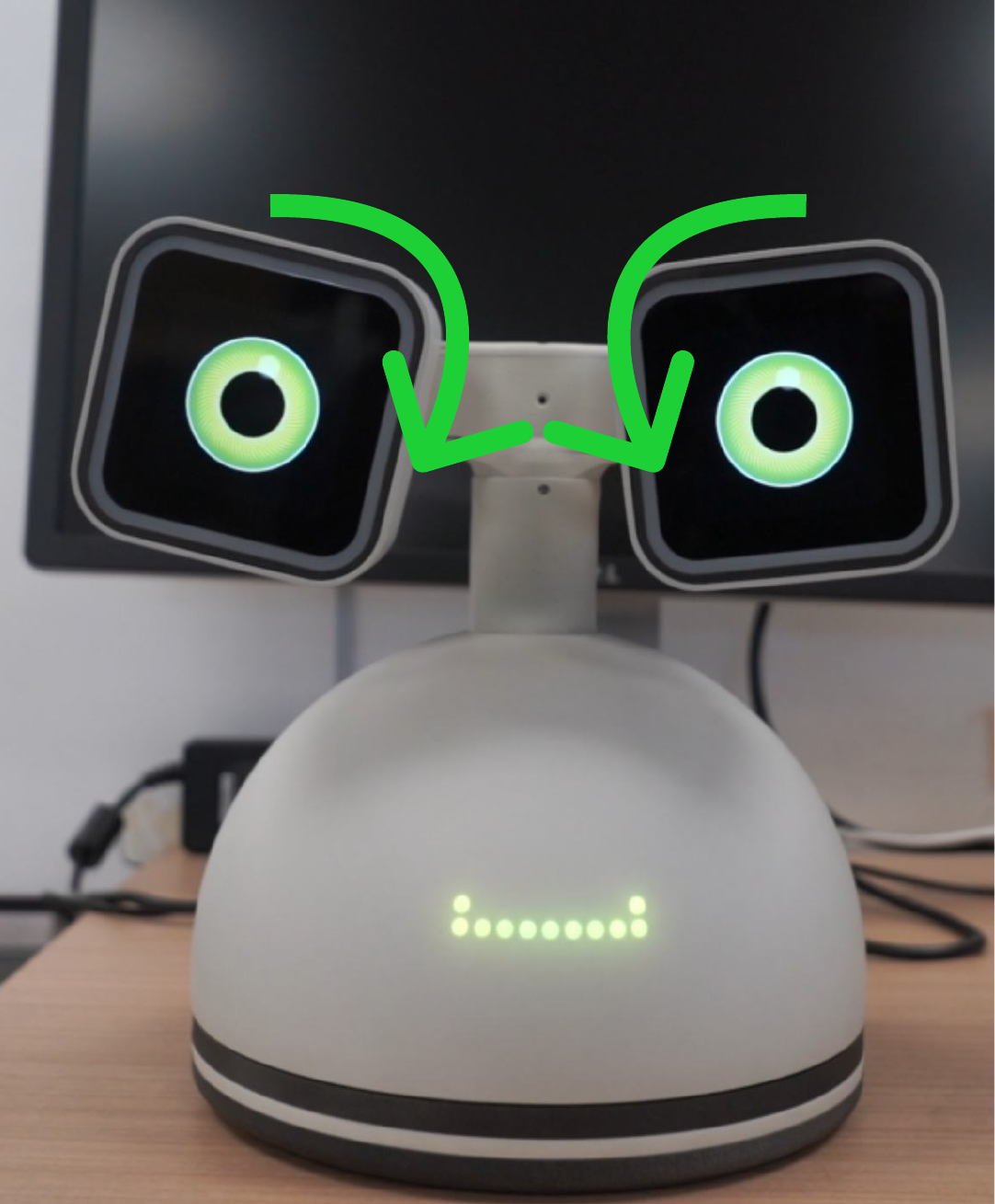} &
    \includegraphics[width=0.15\columnwidth]{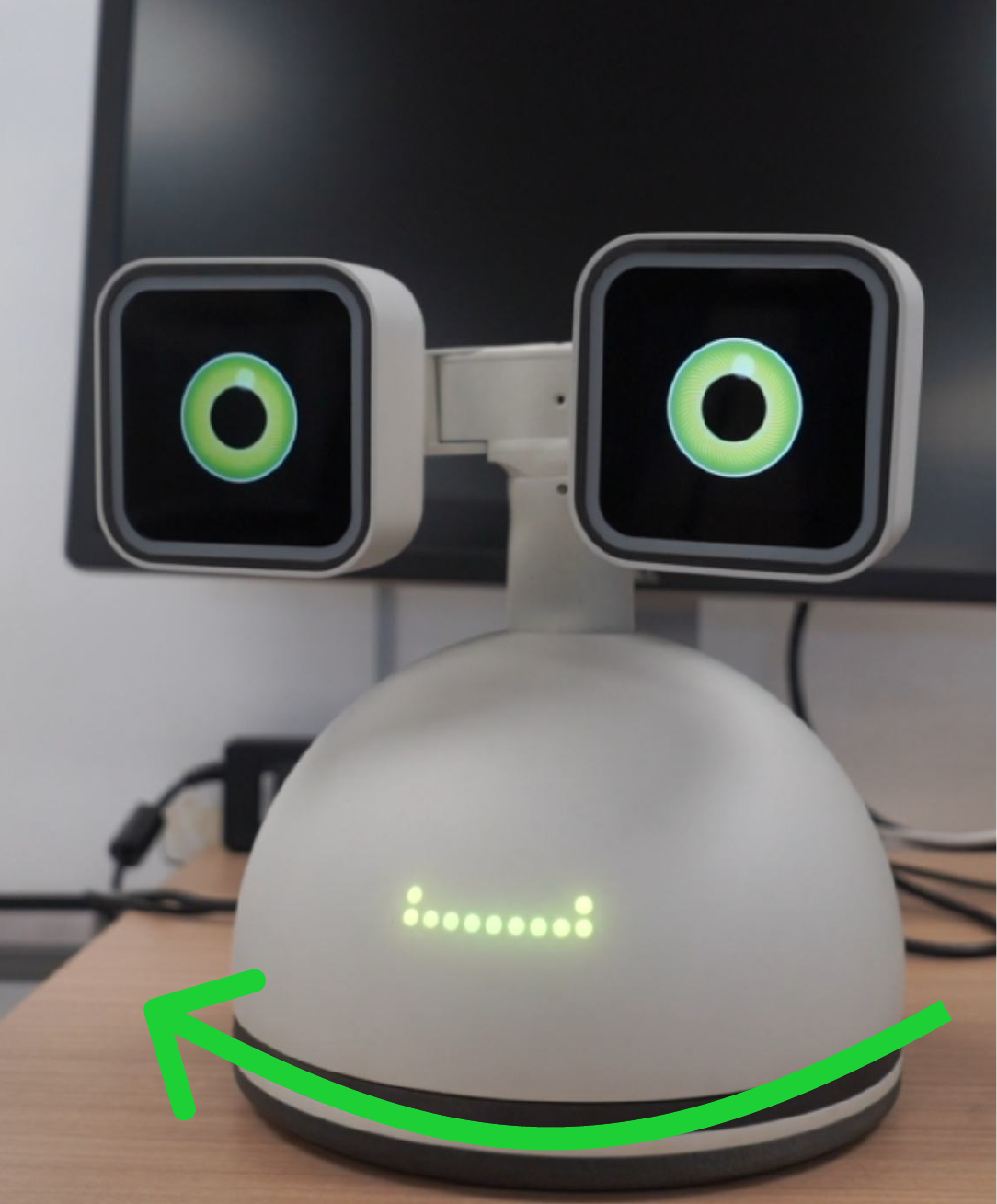} \\
    \bottomrule
  \end{tabular}
\end{table}

\section{Qualitative Evaluation}
\label{sec:evaluation}
We conduct a user study to evaluate the perceived quality of the full system implemented on the Haru robot. Specifically, we measure perceived naturalness of the co-speech gestures and overall preference. To this end, we design a within-subjects study on Prolific, in which participants rate videos of Haru generated by our approach and several baselines.

For the evaluation, we randomly selected two sentences per emotion from our test split (anger, sadness, fear, and joy) from BEAT2, each containing at least one iconic activation. Our institution's ethics committee approved the study, and all participants provided informed consent for their data to be used for research purposes.

\subsection{Baselines}
We compare against several baselines that produce co-speech behaviors under different assumptions. These baselines help us measure overall performance and separate the effect of each component:

\begin{itemize}
    \item \textbf{LLM.} This baseline uses LLM prompts to generate the $Iconic_i$ and $Beat_t$ sequences. We include it because LLMs perform well on many affective computing tasks and they also performed strongly in our quantitative tests (e.g., iconic activation).

    \item \textbf{Ground truth.} This baseline uses the same execution and mapping procedure as our \textit{Robot Gesture Generation} module, but replaces the predicted sequences with the ground-truth BEAT2 labels for iconic gestures and MoCap-derived labels for beats. 

    \item \textbf{Random.} This baseline assigns a random value to all elements of $Iconic_i$. If the activation is $1$, it places a random iconic gesture. Then, it fills the remaining time windows with randomly sampled beat intensities to form $Beat_t$.
\end{itemize}

\subsection{Sample Size and Power Analysis}
We ran an a priori power analysis in G*Power~\cite{faul_gpower_2007} for a repeated-measures ANOVA within factors with four levels. We set $\alpha=0.05$, power $=0.95$, and assumed a medium effect size ($f=0.25$). We also assumed a correlation of $0.5$ between repeated measures and applied a nonsphericity correction of $\epsilon=0.75$. The analysis suggested a minimum sample size of $N=44$. To balance the design across four emotion groups, we included $N=60$ participants ($15$ per emotion).

\subsection{Participants}
We recruited participants through Prolific. We included attention checks to identify workers who did not watch the videos carefully and excluded those who failed. In total, $N=70$ participants completed the study. Of these, $N=60$ passed the attention checks ($30$ female) and we included them in the analysis. Participants received \pounds 6 GBP for approximately $30$ minutes of work.

Participants reported the following age ranges: \textit{18--29} ($N=22$), \textit{30--44} ($N=23$), and \textit{45+} ($N=15$). All participants were native English speakers and reported the following nationalities: UK ($24$), USA ($21$), Canada ($6$), South Africa ($4$), Australia ($3$), New Zealand ($1$), and Nigeria ($1$).

\subsection{Procedure}
Each participant evaluated one emotion condition. We balanced the number of participants across emotions (15 per emotion) and randomized the method order.

Participants first read a short instruction that explained the goal of the study and their task. They then completed a demographics questionnaire and an attention-check question. Next, participants watched four videos showing the same sentences rendered with different methods (our approach and the baselines). After each video, they completed a short questionnaire to evaluate the co-speech behavior and answered additional attention-check questions about the video content.

After rating all videos, participants could re-watch the videos as many times as they wanted and then provided an overall ranking of the methods.

\subsection{Measures}
To assess the effectiveness of the methods, following~\cite{yoon2019robots}, we evaluated three constructs using bipolar items on 5-point Likert scales:

\begin{itemize}
  \item \textbf{Anthropomorphism:} \textit{Fake--Natural}, \textit{Machinelike--Humanlike}, \textit{Unconscious--Conscious}, \textit{Artificial--Lifelike}, \textit{Moving rigidly--Moving elegantly}.
  \item \textbf{Likeability:} \textit{Dislike--Like}, \textit{Unfriendly--Friendly}, \textit{Unkind--Kind}, \textit{Unpleasant--Pleasant}, \textit{Awful--Nice}.
  \item \textbf{Speech--Gesture Correlation:} \textit{Independent--correlated}, \textit{Ignore content--reflect content}, \textit{Not necessary--help to understand content}.
\end{itemize}

Finally, participants ranked the methods by overall quality.

\section{Results} \label{sec:results}
\begin{figure}[tbh]
    \centering
    \includegraphics[width=1.0\linewidth]{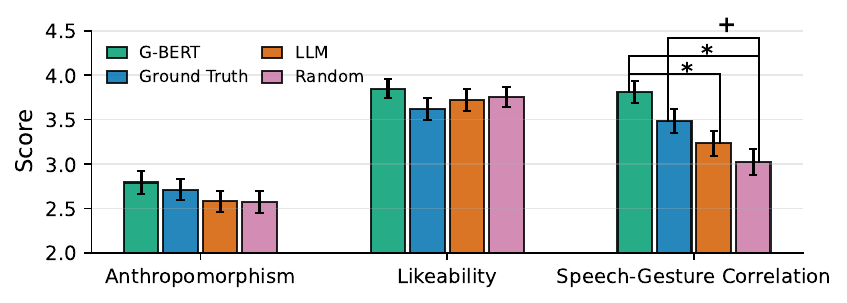}
    \caption{Average score for each questionnaire index across all emotion contexts. Standard error and statistical differences are shown. $+$ : $p < 0.1$, $*$: $p < 0.05$.}
    \label{fig:questions}
\end{figure}

\begin{figure}[tbh]
    \centering
    \includegraphics[width=0.6\linewidth]{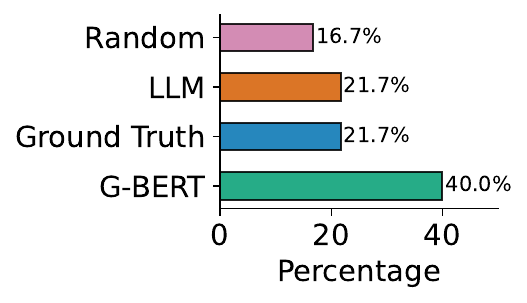}
    \caption{Percentage of trials in which participants ranked each method as the best overall.}
    \label{fig:ranking}
\end{figure}

Figure~\ref{fig:questions} reports the questionnaire results averaged across emotion contexts, and Figure~\ref{fig:ranking} reports the overall preference ranking. We analyze each questionnaire item with a one-way ANOVA across methods. To keep the figures readable, we report significance markers only for items with a significant main effect of method ($p<0.05$). For those, we run a post-hoc method comparison test using Tukey's Honest Significance Difference.

\section{Discussion} \label{sec:discussion}
We observe a significant difference in speech--gesture correlation ($\eta^2$ = 0.0725). G-BERT achieves the highest mean correlation, with significant gains over the LLM and Random baselines and a positive trend over the dataset-label baseline (Figure~\ref{fig:questions}). We attribute this to (i) more accurate iconic timing from BERT, (ii) a beat-intensity distribution from RoBERTa that better matches BEAT2 labels, and (iii) the dataset labels come from human motion, and transferring them to a non-anthropomorphic robot can change how natural the gestures look. Our models adapt timing and intensity to Haru's motion style, which may explain why they can match or exceed the dataset-label baseline in perceived correlation. 

In contrast, we find no significant differences in anthropomorphism or likeability. This may be because these constructs are driven largely by factors held constant across conditions (Haru's embodiment, face, and voice), whereas our methods mainly vary \emph{when} gestures occur and \emph{how strongly} they move. Anthropomorphism remains low for all methods (a likely floor effect for a non-anthropomorphic robot). Likeability shows a non-significant trend in favor of our method, likely due to a ceiling effect and rating variance.

Despite modest quantitative performance on the beat models, our data-driven approach still outperforms the Random and LLM baselines, which suggests it is sufficient for practical applications.

The preference rankings mirror these trends. Participants most often select G-BERT as best (Figure~\ref{fig:ranking}), consistent with its higher speech--gesture correlation scores. We also find that our iconic intensity predictor shifts gestures from predominantly low-intensity in BEAT2 toward more medium- and high-intensity outputs. This may suggest that non-anthropomorphic robots benefit from more salient and frequent iconic gestures.

\section{Limitations and Future Work} \label{sec:limitations}
Our paper has several limitations, which also point to clear next steps. First, our text-only inference improves responsiveness, but it ignores prosody. Future work can add a coarse prosody prediction from text, while keeping latency low. 

Second, our beat module uses a discretized intensity model. In future work, we will explore architectures that capture finer movement variations without sacrificing real-time performance.

Third, our iconic gestures rely on a predefined library; while limiting coverage, curated animations are often needed for non-anthropomorphic robots, and could be augmented with procedural/generative motion.

Fourth, we evaluated via online videos, a valid approach~\cite{hoffman2014designing}. However, future work will test live interactions to assess latency effects.

Fifth, we do not disentangle the effects of iconic and beat motion; future ablations should isolate each module.

Sixth, the system can transfer by defining an embodiment-specific iconic library and specifying the available degrees of freedom and motion ranges for beats, but we evaluated only on Haru; future work should test other platforms.

Finally, since this is the first approximation to a new task, we use separate models for iconic activation, placement/intensity, and beat intensity; future work will explore a unified model to reduce error propagation and improve overall performance.

\section{Conclusions}
We presented an emotion-aware rhythmic--semantic co-speech gesture system that generates iconic and beat behaviors in real time from text. We formalized the task of efficient, emotion-conditioned joint beat--iconic generation and proposed lightweight models for iconic activation/placement/intensity and beat intensity. We integrated the models and deployed the system on the social robot Haru. In an online user study, participants reported stronger speech--gesture alignment and preferred our method over baselines. Overall, our work makes low-latency, emotion-consistent beat--iconic co-speech gesturing practical for social robots.

\bibliographystyle{ieeetr}
\balance
\bibliography{bibliography}
\end{document}